\documentclass[review]{elsarticle}

\usepackage{lineno,hyperref}

\usepackage{url}
\usepackage{graphicx}
\usepackage{bm}
\usepackage{ntheorem}
\usepackage{acronym}
\usepackage{url}
\usepackage{booktabs}
\usepackage{multirow}
\usepackage{float}
\usepackage{xcolor}

\acrodef{LSTM}[LSTM]{Long Short-Term Memory}
\acrodef{ADP}[ADP]{augmented dependency path}
\acrodef{RNN}[RNN]{recursive neural network}
\acrodef{RNN's}[RNN's]{recurrent neural networks}
\acrodef{RE}[RE]{relation extraction}
\acrodef{NLP}[NLP]{natural language processing}
\acrodef{KBs}[KBs]{knowledge bases}
\acrodef{NLP}[NLP]{natural language processing}
\acrodef{CNNs}[CNNs]{Convolutional Neural Networks}
\acrodef{POS}[POS]{Part of Speech}
\acrodef{NERs}[NERs]{named entities recognitions}

\modulolinenumbers[5]

\journal{Applied Soft Computing}

\begin{document}

\begin{frontmatter}

\title{Structural block driven - enhanced convolutional neural representation for relation extraction}
\tnotetext[mytitlenote]{Fully documented templates are available in the elsarticle package on \href{http://www.ctan.org/tex-archive/macros/latex/contrib/elsarticle}{CTAN}.}





\author[add1]{Dongsheng Wang \fnref{fn1}}
\ead{wang@di.ku.dk}

\author[add2]{Prayag Tiwari \fnref{fn1}}
\ead{prayag.tiwari@dei.unipd.it}
\fntext[fn1]{Dongsheng Wang and Prayag Tiwari contribute equally and share the co-first authorship.}

\author[add3]{Sahil Garg}
\ead{sahil.garg@ieee.org}

\author[add5]{Hongyin Zhu}
\ead{zhuhongyin2014@ia.ac.cn}

\author[add4]{Peter Bruza}
\ead{p.bruza@qut.edu.au}

\address[add1]{Department of Computer Science, University of Copenhagen, Copenhagen, Denmark}
\address[add2]{Department of Information Engineering, University of Padova, Padova, Italy}
\address[add3]{\'Ecole de technologie sup\'erieure, Montr\'eal, QC H3C 1K3, Canada}
\address[add5]{Institute of Automation, Chinese Academy of Sciences, Beijing, China}
\address[add4]{School of Information Systems, Queensland University of Technology, 2 George St, Brisbane City, QLD 4000, Australia}

\begin{abstract}
In this paper, we propose a novel lightweight relation extraction approach of structural block driven - convolutional neural learning. Specifically, we detect the essential sequential tokens associated with entities through dependency analysis, named as a structural block, and only encode the block on a block-wise and an inter-block-wise representation, utilizing multi-scale \ac{CNNs}. This is to 1) eliminate the noisy from irrelevant part of a sentence; meanwhile 2) enhance the relevant block representation with both block-wise and inter-block-wise semantically enriched representation. Our method has the advantage of being independent of long sentence context since we only encode the sequential tokens within a block boundary. Experiments on two datasets i.e., SemEval2010 and KBP37, demonstrate the significant advantages of our method. In particular, we achieve the new state-of-the-art performance on the KBP37 dataset; and comparable performance with the state-of-the-art on the SemEval2010 dataset.

\end{abstract}


\end{frontmatter}


\section{Introduction }

Relation extraction (RE) is an essential task in \ac{NLP} to extract the relation between two entities for some given context. In the past years, numerous massive scale \ac{KBs}, including DBpedia \cite{auer2007dbpedia},YAGO \cite{kasneci2006yago}, and  Freebase \cite{bollacker2008freebase}, have been constructed and being broadly utilized in many \ac{NLP} tasks and applications, such as question answering and web search. These knowledge bases mainly consist of relational facts with some format, e.g., (\emph{Google}, \texttt{founder}, \emph{Larry Page}). 

Though existing \ac{KBs} contain a large number of facts, they are not close to encompassing the vast number of facts embedded in plain text. RE is a method of automatically extracting hidden relational facts from plain text to supplement the \ac{KBs}.

The \ac{RE} tasks are considered into two steps, relation detection, and relation classification. Specifically, the first step is to detect candidate relation mentions in sentences involving pairs of entities, and the second is to classify these relation mentions into predefined categories \cite{sebastiani2002machine,pang2002thumbs,kotsiantis2007supervised}. In this paper, we focus on the latter task. 

Non-relation instances can be managed as a normal relation class. 
On the other hand, \ac{RE} comes with a vastly unbalanced dataset where the quantity of non-relation instances far surpasses relation instances, making this \ac{RE} task more challenging but more pragmatic than relation classification. 

In past decades, most of the work in \ac{RE} has been dominated by two approaches which can be differentiated by the essence of the relation description:  kernel-based approaches \cite{qian2008exploiting,zelenko2003kernel,mooney2006subsequence,bunescu2005shortest,zhang2006composite,culotta2004dependency,nguyen2014employing,liu2013convolution}, and feature-based approaches \cite{jiang2007systematic,kambhatla2004combining,nguyen2014employing,chan2010exploiting}. There is a common goal in these approaches which is to leverage a substantial body of knowledge resources and linguistic analysis in order to map relation mentions into some rich representation.
The purpose of the rich representation is that it can be utilized by some statistical classifiers, for example, maximum entropy \cite{nigam1999using,osborne2002using} or support vector machines (SVM) \cite{suykens1999least,scholkopf2001learning}.
The pipeline of linguistic analysis consists of several manually designed steps for example, tokenization, chunking, part of speech tagging, parsing and name tagging, which are often executed by the existing \ac{NLP} module. Due to the knowledge founded by the \ac{NLP} research community, these approaches enable the \ac{RE} module to inherit  knowledge from the pre-processing tasks.
For example, the indicated tasks in the pipeline above are well known to be subject to significant levels of error when applied to out-of-domain dataset \cite{mcclosky2010automatic,jurafsky2006proceedings,sun2016return}, triggering the \ac{RE} module to collapse. Thus, our main aim is to propose a novel \ac{RE} model which reduces the complexity of the feature engineering task, reduce the error propagation rate and improve the performance  in the \ac{RE} task.

In this paper, we propose a novel structural block - driven convolutional neural representation for RE. To be specific, we detect the essential spans associated with entities through dependency relation analysis, by obtaining the parent, siblings, and children nodes of entities.
These are ranked into selective sequential tokens in the same order as they appear in the text. We enhance the selective sequential tokens by enriching them with semantic tags (semantic role and part-of-speech tags), all of which is encoded with multi-scale CNNs. 
Furthermore, we add two more inter-block representations with one subtract layer of the block representation subtracted by the two entity representations, and one multiply layer of the two entity representations. 
Then, we concatenate the block-wise and inter-block wise representations to infer the relation. As a result, the encoding of a selective part of a sentence and the enhanced encoding of the block leads to an improvement in both performance and efficiency. We achieved a new state-of-the-art performance in the KBP37 dataset with an F1 of 60.9, and a comparable result in the SemEval dataset with an F1 of 81.1.

\section{Literature Survey}
Over recent years, many approaches have been proposed for relation extraction and classification. Most related work are based on applying \ac{NLP} system or pattern matching to derive lexical attributes. 
In general, pattern matching is the base for traditional relation classification task \cite{tiwari2019towards,tiwari2018towards} which can be categorized into kernel-based approaches \cite{mooney2006subsequence,bunescu2005shortest} and feature-based approaches \cite{suchanek2006combining,kambhatla2004combining}. The preceding category depends on manually designed patterns and so its time consuming as well as requiring the need the input from experts. Consequently, data sparsity is a challenge facing the latter approaches. Furthermore, extra tools are required for these methods to derive linguistic features.

Distant supervision \cite{angeli2014combining,mintz2009distant,surdeanu2012multi,li2019distant,riedel2010modeling} came into much recognition since 2009 to address the challenge of pattern design, and also because of the scarcity of manuially annotated data. 
This kind of approach integrates knowledge graphs and textual datasets, where the knowledge graph is utilized to  automatically identify patterns from the textual dataset.

 Our approach is inspired by neural models that learn features automatically, e.g.,  Collobert et al \cite{collobert2011natural}. Currently, deep learning is \cite{garg2019hybrid,8758843} very widely applied to learn the underlying feature automatically, so the remainder of the literature survey will cover such approaches. 
 For example, Lin \cite{lin2016neural} proposed a sentence-level attention mechanism to alleviate the wrong labeling problem, expecting to reduce the weight of the instances of noise. They employ a \ac{CNNs} encoder on multiple sentences with selective attention on expected correctly labeled sentences, by adding a learn-able weight to each sentence.
Zhang et. al \citep{zhang2017position} proposed a model that in the first stage amalgamates the \ac{LSTM} sequential approach with the type of entity position-aware attention and shows improvement in relation extraction. In the later stage, TACRED (106,264 instances), a large supervised relation extraction dataset was attained from  crowdsourcing and focused towards TAC KBP relations. 
This combination of an effective model with high quality supervised data yields superior relation extraction performance. The proposed model outperforms all the neural-based baselines. Culotta et. al \cite{culotta2006integrating} proposed a probabilistic extraction model that yields mutual advantage to both "bottom-up" and "top-down" relation extraction. This work demonstrates that amalgamating the relation extraction with pattern discovery improves the performance of each task. 

Zeng et. al \cite{zeng2014relation,zeng2015distant,ji2017distant} used a deep neural network to extract sentence and lexical level features. The proposed architecture takes the input (all the word tokens)  without  complex pre-processing. 
Primarily, all the word tokens are converted to vectors by using word embedding. Furthermore, lexical level based features are extracted according to the stated nouns. Meanwhile, the convolutional model is used to learn the sentence level features. The final extracted feature vector is formed by combining these two-level features. In the end, the obtained features are given to the softmax classifier to predict the association between two marked nouns. The obtained results show that the proposed model outperforms the baselines.  

One interesting work by Nguyen et. al \cite{nguyen2015relation} used \ac{CNNs} for relation extraction that learns features from the sentence automatically and hence reduces the dependencies on external resources and toolkits. The proposed architecture takes benefit of various window sizes for pre-trained word embedding and filter on a non-static architecture as an initializer in order to enhance performance. The relation extraction issues because of unbalanced data have been highlighted in this work. Results shown  improvement, not only over baselines for relation extraction but also relation classification models. 
Liu et. al \cite{liu2015dependency} explored how dependency information can be used. Firstly, the newly termed \ac{ADP} model is proposed, which comprises  the
shortest dependency path among the two subtrees and entities joined to the shortest path. In order to explore the semantic relation behind the \ac{ADP} architecture, they proposed dependency-based neural networks: \ac{CNNs} to apprehend the most essential features on the shortest path, and a \ac{RNN} is constructed to model the subtrees.

Huang et. al \cite{huang2016attention} proposed an attention-based \ac{CNNs} for the relation classification task. The proposed architecture makes full use of word embedding, position embedding and part-of-speech tag embedding information.  which part of the sentence is essential and influential w.r.t the two entities of interest, which is determined by a word level attention approach. 
This model allows learning of some essential features from labeled data, thus removing the dependency on exterior knowledge, for example, the plain dependency structure.  The model was tested on the SemEval-2010
Task 8 benchmark dataset and outperforms all the state-of-art neural network models.
Furthermore, the model can obtain this performance with minimum feature engineering.   

Zhang et. al \cite{zhang2015relation} tried to address the lack of ability to learn temporal features in \ac{CNNs}, and focussing mainly on long-distance dependencies among nominal pairs.  They proposed a general architecture based on \ac{RNN's} and contrasted it with \ac{CNNs}-based approach. A new dataset was introduced which is the refined version of MIMLRE \cite{angeli2014combining}. Experimental results on two datasets demonstrated that \ac{RNN's}-based approach can enhance the performance of relation classification, and, in particular, is proficeint at learning  long distance relationships.

\section{Technical Background}
We introduce the essential technical background of our model, including word embedding, \ac{CNNs}, and semantic parsing. In particular, word embedding initializes the vectors of words; \ac{CNNs} are the representation models that we employ to encode the text; and semantic parsing is the NLP approach we adopted.

\subsection{Multi-scale \ac{CNNs}}

\ac{CNNs} was originally proposed for computer vision but has subsequently been used for text classification \cite{jacovi2018understanding} where it has shown high performance \cite{bai2018empirical,johnson2014effective} which is superior to traditional \ac{NLP} based methods. 

\ac{CNNs} is first utilized in a sentence-level classification by Kim et al. \cite{kim2014convolutional} where they demonstrated improvement on NLP classification tasks.

\begin{figure}
    \centering
     \caption{\label{fig:multi_cnn} Multi-scale \ac{CNNs} \cite{wang2018copenhagen}}
    \includegraphics[width=0.9\textwidth]{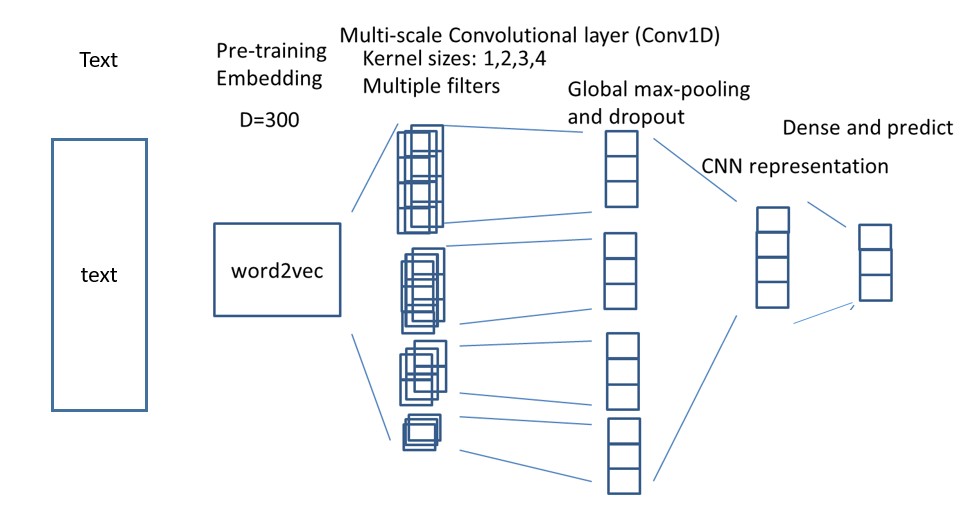}
\end{figure}

Multi-scale \ac{CNNs} have been demonstrated successful \cite{wang2018copenhagen} where they employ multi-scale \ac{CNNs} with different kernel sizes to overcome the drawback of the simple convolutional kernel with fixed window size over encoded semantics of documents, as shown in Figure \ref{fig:multi_cnn}. The reason underlying the design is that determining a fixed window size using a simple convolutional kernel is demanding since small window normally requires deeper networks to gain critical information while large window sizes lead to loss of local information. Hence, multi-scale \ac{CNNs} with multiple window sizes are used to represent the comprehensive contextual information of the text.

When we employ this structure, the last Dense and predict layer will be popped out, resulting in a CNN representation, which can be adopted to encode texts of various types, followed by some new neural encoding scheme.

\subsection{Word Embedding}
Word embedding is a very prominent representation of document vocabulary. This allows capturing the context of some given word in a document, associations with other words, syntactic and semantic similarity, etc. Technically speaking,  there is a mapping of words into vectors containing real number by utilizing the dimension reduction, probabilistic model, or neural networks on some word co-occurrence matrix. This is a kind of feature learning method and language model as well. Word embedding is a kind of process to execute the mapping utilizing neural networks. 

\textbf{Word embedding.} Currently, there are several widely-used word embedding approaches for example, Glove (Stanford) \footnote{\url{https://nlp.stanford.edu/pubs/glove.pdf}}, word2vec (Google)\footnote{\url{https://www.tensorflow.org/tutorials/representation/word2vec}}, and fastest (Facebook)\footnote{\url{https://fasttext.cc/}}. In this work, Glove word embedding is used. Glove\footnote{\url{https://nlp.stanford.edu/projects/glove/}} is type of unsupervised learning model for getting vector representation of words.   

\textbf{Tag Embedding.} Besides, we employ one-hot encoding for POS and dependency tokens in our model, since they are not typical words or terms. Particularly, we encode a size of 24 dimensions embedding for POS labels (24 POS tags in total); and 41 dimensions embedding for dependency labels (41 dependency tags in total) in our model.

\subsection{Semantic processing in NLP}

\subsubsection{Dependency Parsing}
Dependency parsing \footnote{\url{https://nlp.stanford.edu/software/nndep.html}} is the way to investigate the framework of the sentence, forming an association between headwords and also those words which change the heads. The following Figure \ref{fig:dependency} explains the dependency style investigation utilizing the traditional graphical approach which is accepted in the community of dependency parsing. Here it is important to note that the lack of nodes analogous to lexical categories or phrasal constituents in some dependency parse; the inner framework of the dependency parse includes simply the directed associations among lexical components in the sentences. Such  associations directly encode hidden complex phrase structure parses to essential information. For example, arguments of a given verb \emph{prefer} are directly connected in the dependency structure,  although the relation to the main verb is far away in the phrase tree framework. 

\begin{figure}
    \centering
    \caption[Caption for LOF]{\label{fig:dependency} A dependency-style parse alongside the associated constituent-based investigation for \emph {I prefer the morning flight through Denver}\protect\footnotemark}
    \includegraphics[width=0.9\textwidth]{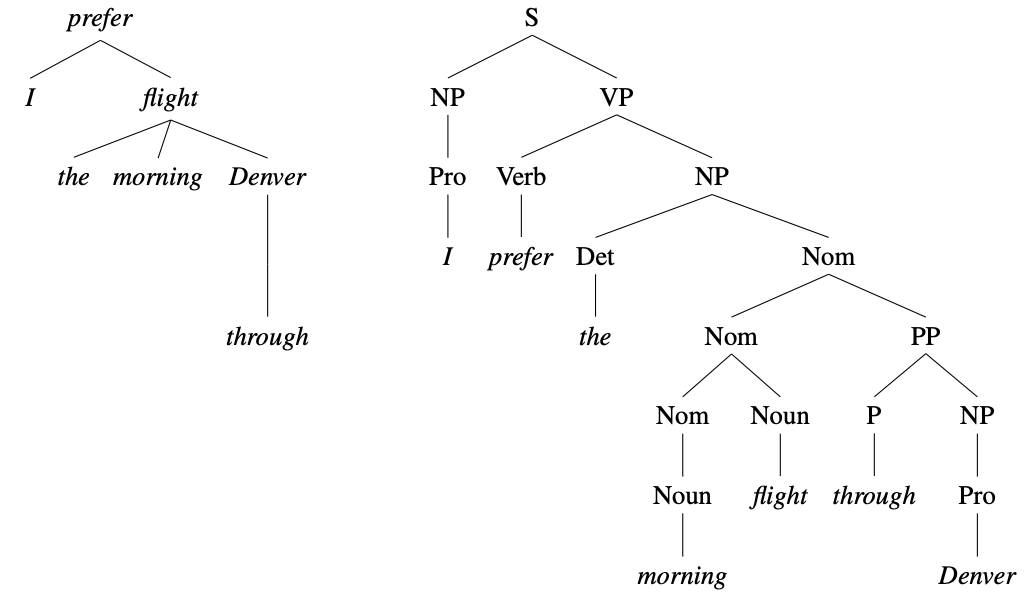}
\end{figure}

\subsubsection{POS Tags}
POS \footnote{\url{https://nlp.stanford.edu/software/tagger.shtml}} tags are very essential for constructing parse tree which are utilized in constructing \ac{NERs} and extract semantic connections among words. POS is also useful for constructing lemmatizers which minimize words to their root form. POS tagging is the way to highlight a word into the documents dataset to an associated POS tag that is based on some context in which then the word is utilized. 

\section{Proposed Method}

We propose a novel approach where the entity oriented structural block is detected, then the semantic representation for the block is encoded utilizing multi-scale \ac{CNNs}, which is further enriched with inter-block representation. We first introduce the structural block detection in section \ref{ss:struct_block}, followed by the enriched semantic encoding using \ac{CNNs} in section \ref{ss:modeling}.

\begin{figure}
    \centering
     \caption{\label{fig:scheme}Our prediction Scheme.}
    \includegraphics[width=0.95\textwidth]{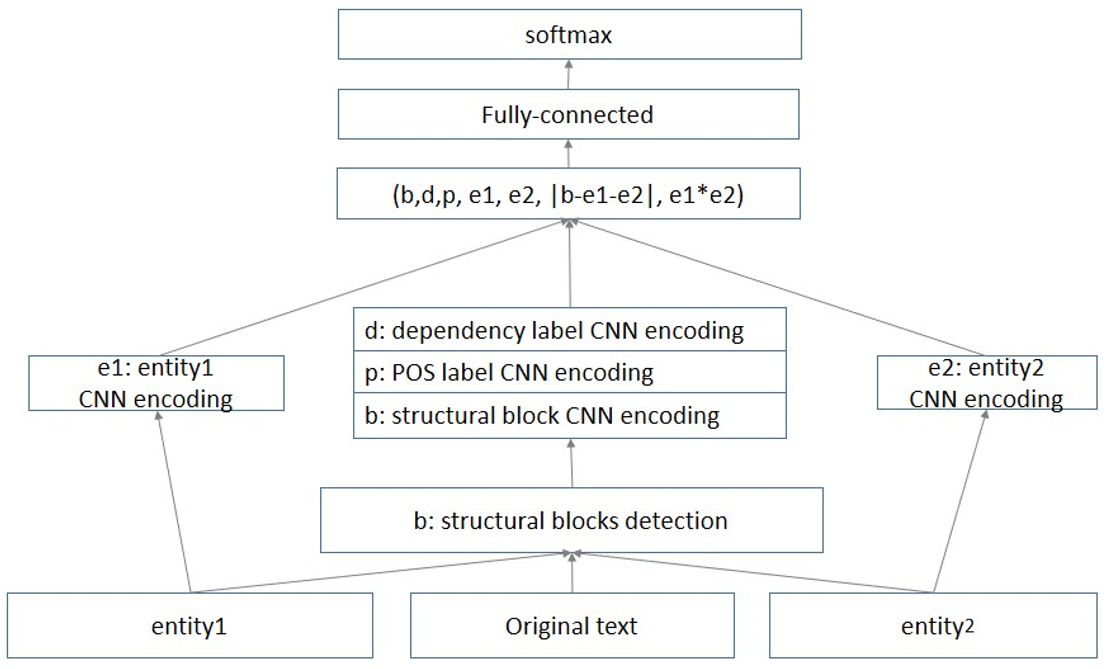}
\end{figure}

\subsection{Structural Block Detection}
\label{ss:struct_block}
We detect the block of coherent tokens for entities in a text. This design is to find directly relevant sequential tokens, whilst retaining their local integrity. First, we obtain the candidate entities, e.g., in the cases of SemEval and KBP37 dataset, they explicitly give two entities. Next, we detect their structurally related tokens to restore their coherent structural semantics. 
We achieve this by building up a dependency tree for each sentence with dependency relations between tokens and find the parent, siblings and children nodes as a single block for each entity $e_i$, which is defined as Eq. \ref{eq:bloc}. 

\begin{equation}
\label{eq:bloc}
single\_block(e_i) = \sum_{t^j \in e_i} {(t^j+head(t^j)+siblings(t^j)+children(t^j))}
\end{equation}

where $t^j$ is a token from the entity with its rank position $j$ in text; $head(t^j)$ basically refers to the relation name (mostly verbs) while $siblings(t^j)$ refers to those tokens that share the same relation name; and the $children(t^j)$ indicates those tokens that depend on $t^j$. A single block literally covers all the structurally related tokens in terms of a single entity. Nevertheless, we provide an alternative version without including children, i.e., $single\_block(e_i) = \sum_{t^j \in e_i} {(t^j+head(t^j)+siblings(t^j))}$. This version empirically leads to comparable or better performance than the first version, the comparison of which is detailed in section \ref{ss:result}. 

The selected tokens for a set of entity $E$ is defined as Eq. \ref{eq:tokens} where all single blocks are aggregated into one block defined as $aggreg\_block(E)$. All the token indexes that are selected are denoted as $J=set(\sum_{t^j \in aggreg\_block(E)} j)$, indicating that duplicate j will be removed.

\begin{equation}
\label{eq:tokens}
aggreg\_block(E) = \sum_{e_i \in E} single\_block(e_i)
\end{equation}


Then, the sequential tokens for the block are defined in Eq. \ref{eq:seq_tok}.

\begin{equation}
\label{eq:seq_tok}
seq\_tokens(E) = \bigoplus_{j \in ranked(J)}^{|J|} t^j 
\end{equation}
 
As the block is locally integrated instead of being integrated through the whole sentence, we additionally concatenate its semantic role $role(t)$ and the part-of-speech $pos(t)$ to assist the generalized learning on the selective sequential tokens. The resulting structural block for a sentence is defined as Eq.\ref{eq:struct_block},
\begin{equation}
\label{eq:struct_block}
structural\_block(E) = \sum_{t \in seq\_tokens(E)} (t \oplus role(t) \oplus pos(t))
\end{equation}
where $structural\_block(E)$ is the enhanced representation for the selective structural block.

As a result, the relation predicting is defined with softmax function of the probability distribution as below,
\begin{equation}
    p(r|\theta,s) = softmax(M(h\odot s)+b) 
\end{equation}

where $softmax(x)=\frac{Exp(x)}{\sum_{k=1}^K Exp(x_k)}$ ($K$ is the number of the relation); $M$ is the matrix representation of the relation, $s$ is the enhanced structural block representation; $h$ is the hidden layer; and $b$ is a bias vector in terms of the output.




\subsection{Modeling}
\label{ss:modeling}
Inspired by \cite{conneau2017supervised}, we propose an adapted model scheme as shown in Figure \ref{fig:scheme}. The design is to take full advantage of the selective structural blocks with both block and inter-block representation. 

For a given sentence and two entities, we first detect the structural block with dependency analysis, resulting in a subset of sequential tokens. We then enrich the semantics by concatenating their semantic role tags and POS tags as demonstrated in Eq.  \ref{eq:struct_block}. As a result, every single token of the block is represented with three tokens, working together to represent the whole structural block. The semantic role tokens are supposed to give rise to the generalized learning on the selective sequential tokens. 

Then, we encode these sequential tokens with multi-scale \ac{CNNs}, as shown in Figure \ref{fig:multi_cnn}. Specifically, we pop out the softmax layer of the model, resulting in the previous CNN representation layer, to encode the sequential tokens of the entities, dependency labels, POS labels, and structural block. 

Besides, we explicitly encode the inter-block representation to gain the connection among the block and two entities. The inter-block representation includes a subtract layer between the block and the two entities, expressed as $b-e1-e2$; and a multiply layer between two entities expressed as $e1*e2$, assisting the similarity inference between the two entities (as $sim(e1,e2) = \frac{e1* e2}{|e1|.|e2|} $). 

As a result, we fully connect the structural block representation and the inter-block representation and softmax the final relation classification. 

\section{Experimental Results}

\subsection{Data}
Semantic Evaluation \footnote{\url{https://www.cs.york.ac.uk/semeval2010_WSI/datasets.html}} dataset has 10 relations, as listed below. However, because they have order difference, therefore, there is a total of 19 relations, as the "Other" does not apply this rule.

The second KBP37 dataset \footnote{\url{https://github.com/davidsbatista/Annotated-Semantic-Relationships-Datasets/issues/3}} is the revised version of MIML-RE annotation dataset, which was provided by Gabor Angeli et al. \cite{angeli2014combining}. They utilized both the 2013 and 2010 KBP document dataset, as well as a July 2013 dump of Wikipedia as the text dataset for further annotation. There are 33811 annotated sentences. More details about this dataset are presented in table \ref{tb:data_statistics} and \ref{tb:desc_data}.


%
\begin{table}[]
\caption{\label{tb:data_statistics} Dataset statistics.}
\centering
\begin{tabular}{@{}lll@{}}
\toprule
Dataset & train (instances) & test (instances) \\ \midrule
SemEval2010 & 8,000 & 2,717 \\
KBP37 & 15,917 & 3,405 \\ \bottomrule
\end{tabular}
\end{table}

\begin{table}[!htb]
\caption{\label{tb:desc_data} Details about SemEval2010 and KBP37 datasets.}
\centering
\small\addtolength{\tabcolsep}{-5pt}
\begin{tabular}{|l|l|l|l|}
\hline
\multirow{7}{*}{SemEval2010} & No. of relation types: & \multicolumn{2}{l|}{19} \\ \cline{2-4} 
 & No. of relation classes: & \multicolumn{2}{l|}{10} \\ \cline{2-4} 
 & \multirow{5}{*}{Classes:} & Cause-Effect & Instrument-Agency \\ \cline{3-4} 
 &  & Product-Producer & Content-Container \\ \cline{3-4} 
 &  & Entity-Origin & Entity-Destination \\ \cline{3-4} 
 &  & Component-Whole & Member-Collection \\ \cline{3-4} 
 &  & Communication-Topic & Other \\ \hline
\multirow{11}{*}{KBP37} & No. of relation types: & \multicolumn{2}{l|}{37} \\ \cline{2-4} 
 & No. of relation classes: & \multicolumn{2}{l|}{19} \\ \cline{2-4} 
 & \multirow{9}{*}{Classes:} & per:alternate\_names & org:alternate\_names \\ \cline{3-4} 
 &  & per:origin & org:subsidiaries \\ \cline{3-4} 
 &  & per:spouse & org:top\_members/employees \\ \cline{3-4} 
 &  & per:title & org:founded \\ \cline{3-4} 
 &  & per:employee|\_of & org:founded\_by \\ \cline{3-4} 
 &  & per:countries\_of\_residence & org:countries\_of\_headquarters \\ \cline{3-4} 
 &  & per:stateorprovince\_of\_residence & org:stateorprovince\_of\_headquarters \\ \cline{3-4} 
 &  & per:country\_of\_birth & org:member \\ \cline{3-4} 
 &  & no\_relation &  \\ \hline
\end{tabular}
\end{table}

\subsection{Hardware setting}
We list the hardware settings used to conduct the experiments. It is importrant to note that we employ ordinary CPU server settings instead of GPUs.
Table \ref{tb:hardware_setting} indicates that our model scheme overall has low setting requirements comparable to most existing servers. We will demonstrate that even on such basic servers the training speed is both fast and efficient.

\begin{table}[]
\caption{\label{tb:hardware_setting} Hardware settings.}
\begin{tabular}{@{}llllll@{}}
\toprule
\textbf{Property} & \textbf{modes} & \textbf{CPU} & \textbf{Memory} & \textbf{System} & \textbf{Threads per core} \\ \midrule
\textbf{Value} & 64 bits & 40 & 125G & Ubuntu 14.04 & 2 \\ \bottomrule
\end{tabular}
\end{table}

\subsection{Evaluation Measures}
There are several evaluation measures in machine learning for example, accuracy, precision, recall, f-score, etc. In line with most related work, we employ the F-score to evaluate the performance of our model and use it to compare the performance with other state-of-art neural models.

\begin{equation}
\label{eq:precision}
Precision (P) = \frac{Number \; of \; correctly  \; extracted \; entity \; relations}{Total \;number \;of \;extracted \;entity \;relations}
\end{equation}

\begin{equation}
\label{eq:recall}
Recall (R) =
\frac{Number \;of \;correctly \;extracted \;entity \;relations}{Actual \;number \;of \;extracted \;entity \;relations}
\end{equation}

\begin{equation}
\label{eq:f1}
F1 = \frac{2P*R}{P+R}
\end{equation}

\subsection{Results and Discussion}
\label{ss:result}
\begin{figure}[!hbt]
    \centering
     \caption{\label{fig:kbp_cm} KBP37 confusion matrix, label mappings is shown at the bottom of the result section \ref{ss:result}.}
    \includegraphics[width=1.0\textwidth]{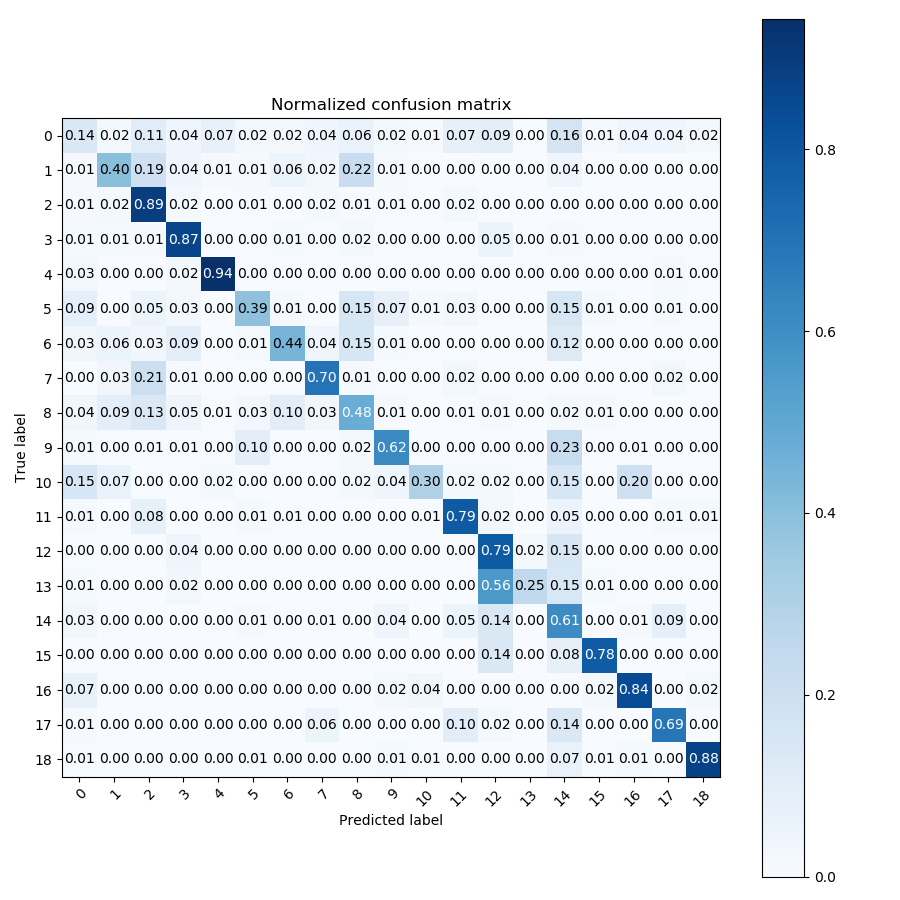}
\end{figure}

As shown in Table \ref{tb:result_overall}, our model has superior performance on the KBP37 dataset and comparable performance on the SemEval2010 dataset. The main advantage of our method is that we do not rely on manually determined features while all the other methods adopt a feature set to some degree. As we achieve the new state-of-art performance in KBP37 (with an improvement of 2.1\%), we found that text context in KBP37 is much longer than in SemEval2010.
Consequently, it seems  that our method has a superior advantage in long contexts, since the selective structural block strongly reduces the higher levels of noise associated with long contexts.
The confusion matrix is shown in Figure \ref{fig:kbp_cm}, where the mappings of label is listed at footnote \footnote{$no\_relation=0; org\_alternate\_names= 1; org\_city\_of\_headquarters= 2; org\_country\_of\_headquarters= 3; org\_founded= 4; org\_founded\_by= 5; org\_members= 6; org\_stateorprovince\_of\_headquarters= 7; org\_subsidiaries= 8; org\_top\_members= 9; per\_alternate\_names= 10; per\_cities\_of\_residence= 11; per\_countries\_of\_residence= 12; per\_country\_of\_birth= 13; per\_employee\_of= 14; per\_origin= 15; per\_spouse= 16; per\_stateorprovinces\_of\_residence= 17; per\_title= 18$}.

\begin{figure}[!hbt]
    \centering
     \caption{\label{fig:semeval_cm} SemEval2010 confusion matrix}
    \includegraphics[width=1.0\textwidth]{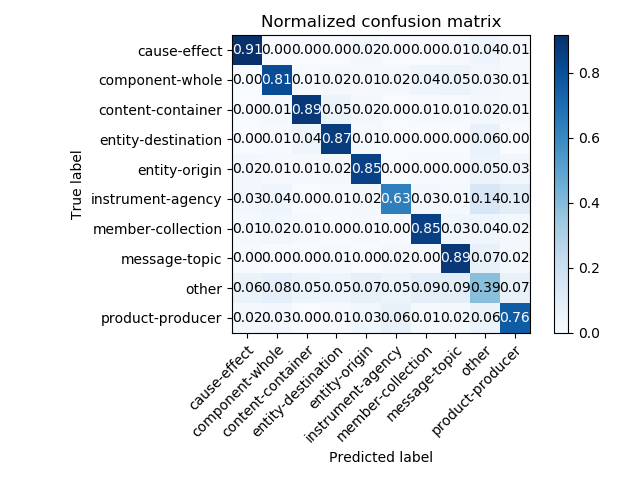}
\end{figure}

In SemEval2010, we found that the Instrument-Agency has the lowest F1 (below 70\%), which is the bottleneck for the overall results. The reason behind this category can be that it has broader coverage of instances for the two types, namely, Instrument and Agency; while as we can observe that Cause-Effect, Member-Collection, etc. have more specific patterns, or smaller coverage of instances. There are some state-of-art models \cite{zhou2016attention,zhang2018graph} whose F-score is slightly higher than our  model, but our model take less time to compute than others.
Further, our model rely on manual determined features while all the other methods adopt some level of manually selected features. The confusion matrix is shown in Figure \ref{fig:semeval_cm}. 

Our method exhibits satisfactory training speed as well. The average epoch training time is 4.6s on SemEval2010 and 6.3s on KBP37 datasets, due to \ac{CNNs} and the structrual block detection. Though we are not provided with the training time from other published models for comparison, we claim the satisfactory efficiency of our method.

As discussed in Eq. \ref{eq:bloc}, we empirically found that the adoption of children nodes when detecting block does not necessarily influence the final performance. As shown in table \ref{tb:compare_child}, the performance of with-children version is slightly lower than that of without-children; while it is the opposite for KBP37 dataset. Therefore, we maintain that the single block detection can include or exclude the children nodes.

A single model shows an F1 between 78\% to 79\% in SemEval2010 dataset, but  improves to 81.1\% when we ensemble the results from 2-4 models that are trained from the same scheme. Given the proposed scheme is independent of the fundamental encoding model, we therefore assume that different representation models, including LSTM, BiLSTM, or a mixture modeling can replace our multi-\ac{CNNs}; and an ensemble of models with different fundamental representations can potentially improve the ultimate performance. 

\begin{table}[!hbt]
\caption{\label{tb:compare_child} Comparison of two versions of obtaining blocks.}
\centering
\small
\begin{tabular}{@{}|l|l|l|@{}}
\toprule
Our Model Version & SemEval2010 & KBP37 \\ \midrule
 & \multicolumn{2}{l|}{Macro F1 (\%)} \\ \midrule
Block with-children & 80.7 & 60.9 \\ \midrule
Block without-children & 81.1 & 60.7 \\ \bottomrule
\end{tabular}
\end{table}

\begin{table}[!hbt]
\caption{\label{tb:result_overall} Results of different neural models compared to our proposed models}
\centering
\small
\begin{tabular}{@{}|c|c|c|@{}}
\toprule
\multirow{2}{*}{Model} & Semeval-2010 & KBP37 \\ \cmidrule(l){2-3} 
 & \multicolumn{2}{l|}{  Macro F1 (\%)} \\ \midrule

MV-RNN \cite{socher2012semantic}  & 79.1 & - \\ \midrule
CNN+PF \cite{zeng2014relation} & 78.9 & - \\ \midrule
CNN+PF  \cite{zhang2015relation} & 78.3 & 51.3 \\ \midrule
CNN+PI \cite{zhang2015relation} & 77.4 & 55.1 \\ \midrule
RNN+PF  \cite{zhang2016relation} & 78.8 & 54.3 \\ \midrule
RNN+PI \cite{zhang2016relation}& 79.6 & 58.8 \\ \midrule
att-BLSTM \cite{zhou2016attention}& 84.0 & - \\ \midrule
GCN \cite{zhang2018graph} & 84.8 & - \\ \midrule
Our Model & \textbf{81.1} & \textbf{60.9} \\ \bottomrule
\end{tabular}
\end{table}

\begin{figure}[!h]
    \centering
     \caption{\label{fig:bar}F1-score on Semeval-2010 and KBP37}
    \includegraphics[width=14cm, ]{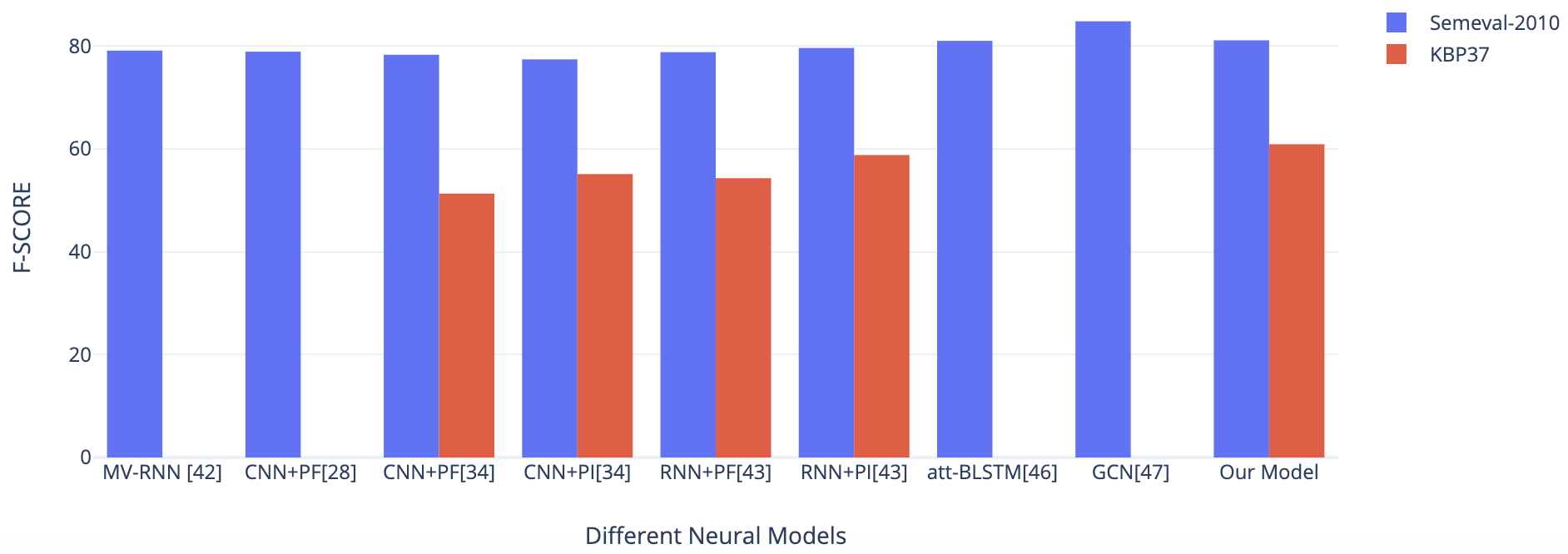}
\end{figure}

\section{Conclusion and Future Work}

Relation extraction plays an important role in the population of KBs and other NLP tasks. In this paper, we presented a novel relation extraction approach with few features called the structural block - driven convolutional neural model. The design of the model is to 1) eliminate the noise due to irrelevant parts of a sentence and 2) enhance the relevant block representation, by adopting semantic role embedding and concatenating the inter-block representation. The block is detected by entity oriented dependency analysis, and the enhanced encoding of the block is conducted not only on block-wise representation but with two more layers, one subtract and one multiply layer, based on strong multi-scale convolutional neural models.

We validated our model on two datasets, i.e., SemEval2010 and KBP37, where we achieve the new state-of-the-art performance on the KBP37 dataset (60.9 F1) and  comparable performance with the state-of-the-art on the SemEval2010 dataset (81.1 F1). 

Our method has a superior advantage in long sentence contexts since our model only encodes  sequential tokens within the block boundary. For example, the performance in KBP37 demonstrated the model's robustness in long contexts. Moreover, compared to most of the other relation extraction approaches, we do not rely on a manually constructed feature set. Finally, the model has a satisfactory training speed, e.g., 4.6s epoch training time in SemEval2010 and 6.3s in KBP37. 

As the scheme we proposed is extensible, one avenue for future work is to apply the scheme based on other basic encoders, instead of CNNs. For instance, BERT \cite{devlin2018bert} has proved to be highly efficient in multiple NLP tasks. We can replace the CNNs with  BERT as an enhanced BERT representation to observe whether it can generate further improvement.


\section{Acknowledgements}

This project receive funding from the European Union's Horizon 2020 research and innovation programme under the Marie Sklodowska-Curie grant agreement No 721321".

\bibliography{mybibfile}

\begin{thebibliography}{10}
\expandafter\ifx\csname url\endcsname\relax
  \def\url#1{\texttt{#1}}\fi
\expandafter\ifx\csname urlprefix\endcsname\relax\def\urlprefix{URL }\fi
\expandafter\ifx\csname href\endcsname\relax
  \def\href#1#2{#2} \def\path#1{#1}\fi

\bibitem{auer2007dbpedia}
S.~Auer, C.~Bizer, G.~Kobilarov, J.~Lehmann, R.~Cyganiak, Z.~Ives, Dbpedia: A
  nucleus for a web of open data, in: The semantic web, Springer, 2007, pp.
  722--735.

\bibitem{kasneci2006yago}
G.~Kasneci, F.~Suchanek, G.~Weikum, Yago-a core of semantic knowledge.

\bibitem{bollacker2008freebase}
K.~Bollacker, C.~Evans, P.~Paritosh, T.~Sturge, J.~Taylor, Freebase: a
  collaboratively created graph database for structuring human knowledge, in:
  Proceedings of the 2008 ACM SIGMOD international conference on Management of
  data, AcM, 2008, pp. 1247--1250.

\bibitem{sebastiani2002machine}
F.~Sebastiani, Machine learning in automated text categorization, ACM computing
  surveys (CSUR) 34~(1) (2002) 1--47.

\bibitem{pang2002thumbs}
B.~Pang, L.~Lee, S.~Vaithyanathan, Thumbs up?: sentiment classification using
  machine learning techniques, in: Proceedings of the ACL-02 conference on
  Empirical methods in natural language processing-Volume 10, Association for
  Computational Linguistics, 2002, pp. 79--86.

\bibitem{kotsiantis2007supervised}
S.~B. Kotsiantis, I.~Zaharakis, P.~Pintelas, Supervised machine learning: A
  review of classification techniques, Emerging artificial intelligence
  applications in computer engineering 160 (2007) 3--24.

\bibitem{qian2008exploiting}
L.~Qian, G.~Zhou, F.~Kong, Q.~Zhu, P.~Qian, Exploiting constituent dependencies
  for tree kernel-based semantic relation extraction, in: Proceedings of the
  22nd International Conference on Computational Linguistics-Volume 1,
  Association for Computational Linguistics, 2008, pp. 697--704.

\bibitem{zelenko2003kernel}
D.~Zelenko, C.~Aone, A.~Richardella, Kernel methods for relation extraction,
  Journal of machine learning research 3~(Feb) (2003) 1083--1106.

\bibitem{mooney2006subsequence}
R.~J. Mooney, R.~C. Bunescu, Subsequence kernels for relation extraction, in:
  Advances in neural information processing systems, 2006, pp. 171--178.

\bibitem{bunescu2005shortest}
R.~C. Bunescu, R.~J. Mooney, A shortest path dependency kernel for relation
  extraction, in: Proceedings of the conference on human language technology
  and empirical methods in natural language processing, Association for
  Computational Linguistics, 2005, pp. 724--731.

\bibitem{zhang2006composite}
M.~Zhang, J.~Zhang, J.~Su, G.~Zhou, A composite kernel to extract relations
  between entities with both flat and structured features, in: Proceedings of
  the 21st International Conference on Computational Linguistics and the 44th
  annual meeting of the Association for Computational Linguistics, Association
  for Computational Linguistics, 2006, pp. 825--832.

\bibitem{culotta2004dependency}
A.~Culotta, J.~Sorensen, Dependency tree kernels for relation extraction, in:
  Proceedings of the 42nd annual meeting on association for computational
  linguistics, Association for Computational Linguistics, 2004, p. 423.

\bibitem{nguyen2014employing}
T.~H. Nguyen, R.~Grishman, Employing word representations and regularization
  for domain adaptation of relation extraction, in: Proceedings of the 52nd
  Annual Meeting of the Association for Computational Linguistics (Volume 2:
  Short Papers), 2014, pp. 68--74.

\bibitem{liu2013convolution}
C.~Liu, W.~Sun, W.~Chao, W.~Che, Convolution neural network for relation
  extraction, in: International Conference on Advanced Data Mining and
  Applications, Springer, 2013, pp. 231--242.

\bibitem{jiang2007systematic}
J.~Jiang, C.~Zhai, A systematic exploration of the feature space for relation
  extraction, in: Human Language Technologies 2007: The Conference of the North
  American Chapter of the Association for Computational Linguistics;
  Proceedings of the Main Conference, 2007, pp. 113--120.

\bibitem{kambhatla2004combining}
N.~Kambhatla, Combining lexical, syntactic, and semantic features with maximum
  entropy models for information extraction, in: Proceedings of the ACL
  Interactive Poster and Demonstration Sessions, 2004, pp. 178--181.

\bibitem{chan2010exploiting}
Y.~S. Chan, D.~Roth, Exploiting background knowledge for relation extraction,
  in: Proceedings of the 23rd International Conference on Computational
  Linguistics, Association for Computational Linguistics, 2010, pp. 152--160.

\bibitem{nigam1999using}
K.~Nigam, J.~Lafferty, A.~McCallum, Using maximum entropy for text
  classification, in: IJCAI-99 workshop on machine learning for information
  filtering, Vol.~1, 1999, pp. 61--67.

\bibitem{osborne2002using}
M.~Osborne, Using maximum entropy for sentence extraction, in: Proceedings of
  the ACL-02 Workshop on Automatic Summarization-Volume 4, Association for
  Computational Linguistics, 2002, pp. 1--8.

\bibitem{suykens1999least}
J.~A. Suykens, J.~Vandewalle, Least squares support vector machine classifiers,
  Neural processing letters 9~(3) (1999) 293--300.

\bibitem{scholkopf2001learning}
B.~Scholkopf, A.~J. Smola, Learning with kernels: support vector machines,
  regularization, optimization, and beyond, MIT press, 2001.

\bibitem{mcclosky2010automatic}
D.~McClosky, E.~Charniak, M.~Johnson, Automatic domain adaptation for parsing,
  in: Human Language Technologies: The 2010 Annual Conference of the North
  American Chapter of the Association for Computational Linguistics,
  Association for Computational Linguistics, 2010, pp. 28--36.

\bibitem{jurafsky2006proceedings}
D.~Jurafsky, E.~Gaussier, Proceedings of the 2006 conference on empirical
  methods in natural language processing, in: Proceedings of the 2006
  Conference on Empirical Methods in Natural Language Processing, 2006.

\bibitem{sun2016return}
B.~Sun, J.~Feng, K.~Saenko, Return of frustratin2gly easy domain adaptation,
  in: Thirtieth AAAI Conference on Artificial Intelligence, 2016.

\bibitem{tiwari2019towards}
P.~Tiwari, M.~Melucci, Towards a quantum-inspired binary classifier, IEEE
  Access 7 (2019) 42354--42372.

\bibitem{tiwari2018towards}
P.~Tiwari, M.~Melucci, Towards a quantum-inspired framework for binary
  classification, in: Proceedings of the 27th ACM International Conference on
  Information and Knowledge Management, ACM, 2018, pp. 1815--1818.

\bibitem{suchanek2006combining}
F.~M. Suchanek, G.~Ifrim, G.~Weikum, Combining linguistic and statistical
  analysis to extract relations from web documents, in: Proceedings of the 12th
  ACM SIGKDD international conference on Knowledge discovery and data mining,
  ACM, 2006, pp. 712--717.

\bibitem{angeli2014combining}
G.~Angeli, J.~Tibshirani, J.~Wu, C.~D. Manning, Combining distant and partial
  supervision for relation extraction, in: Proceedings of the 2014 conference
  on empirical methods in natural language processing (EMNLP), 2014, pp.
  1556--1567.

\bibitem{mintz2009distant}
M.~Mintz, S.~Bills, R.~Snow, D.~Jurafsky, Distant supervision for relation
  extraction without labeled data, in: Proceedings of the Joint Conference of
  the 47th Annual Meeting of the ACL and the 4th International Joint Conference
  on Natural Language Processing of the AFNLP: Volume 2-Volume 2, Association
  for Computational Linguistics, 2009, pp. 1003--1011.

\bibitem{surdeanu2012multi}
M.~Surdeanu, J.~Tibshirani, R.~Nallapati, C.~D. Manning, Multi-instance
  multi-label learning for relation extraction, in: Proceedings of the 2012
  joint conference on empirical methods in natural language processing and
  computational natural language learning, Association for Computational
  Linguistics, 2012, pp. 455--465.

\bibitem{li2019distant}
J.~Li, R.~Hu, X.~Liu, P.~Tiwari, H.~M. Pandey, W.~Chen, B.~Wang, Y.~Jin,
  K.~Yang, A distant supervision method based on paradigmatic relations for
  learning word embeddings, Neural Computing and Applications  1--10.

\bibitem{riedel2010modeling}
S.~Riedel, L.~Yao, A.~McCallum, Modeling relations and their mentions without
  labeled text, in: Joint European Conference on Machine Learning and Knowledge
  Discovery in Databases, Springer, 2010, pp. 148--163.

\bibitem{collobert2011natural}
R.~Collobert, J.~Weston, L.~Bottou, M.~Karlen, K.~Kavukcuoglu, P.~Kuksa,
  Natural language processing (almost) from scratch, Journal of machine
  learning research 12~(Aug) (2011) 2493--2537.

\bibitem{garg2019hybrid}
S.~Garg, K.~Kaur, N.~Kumar, J.~J. Rodrigues, Hybrid deep-learning-based anomaly
  detection scheme for suspicious flow detection in sdn: A social multimedia
  perspective, IEEE Transactions on Multimedia 21~(3) (2019) 566--578.

\bibitem{8758843}
S.~{Garg}, K.~{Kaur}, N.~{Kumar}, G.~{Kaddoum}, A.~Y. {Zomaya}, R.~{Ranjan}, A
  hybrid deep learning-based model for anomaly detection in cloud datacenter
  networks, IEEE Transactions on Network and Service Management 16~(3) (2019)
  924--935.
\newblock \href {http://dx.doi.org/10.1109/TNSM.2019.2927886}
  {\path{doi:10.1109/TNSM.2019.2927886}}.

\bibitem{lin2016neural}
Y.~Lin, S.~Shen, Z.~Liu, H.~Luan, M.~Sun, Neural relation extraction with
  selective attention over instances, in: Proceedings of the 54th Annual
  Meeting of the Association for Computational Linguistics (Volume 1: Long
  Papers), Vol.~1, 2016, pp. 2124--2133.

\bibitem{zhang2017position}
Y.~Zhang, V.~Zhong, D.~Chen, G.~Angeli, C.~D. Manning, Position-aware attention
  and supervised data improve slot filling, in: Proceedings of the 2017
  Conference on Empirical Methods in Natural Language Processing, 2017, pp.
  35--45.

\bibitem{culotta2006integrating}
A.~Culotta, A.~McCallum, J.~Betz, Integrating probabilistic extraction models
  and data mining to discover relations and patterns in text, in: Proceedings
  of the main conference on Human Language Technology Conference of the North
  American Chapter of the Association of Computational Linguistics, Association
  for Computational Linguistics, 2006, pp. 296--303.

\bibitem{zeng2014relation}
D.~Zeng, K.~Liu, S.~Lai, G.~Zhou, J.~Zhao, et~al., Relation classification via
  convolutional deep neural network.

\bibitem{zeng2015distant}
D.~Zeng, K.~Liu, Y.~Chen, J.~Zhao, Distant supervision for relation extraction
  via piecewise convolutional neural networks, in: Proceedings of the 2015
  Conference on Empirical Methods in Natural Language Processing, 2015, pp.
  1753--1762.

\bibitem{ji2017distant}
G.~Ji, K.~Liu, S.~He, J.~Zhao, Distant supervision for relation extraction with
  sentence-level attention and entity descriptions, in: Thirty-First AAAI
  Conference on Artificial Intelligence, 2017.

\bibitem{nguyen2015relation}
T.~H. Nguyen, R.~Grishman, Relation extraction: Perspective from convolutional
  neural networks, in: Proceedings of the 1st Workshop on Vector Space Modeling
  for Natural Language Processing, 2015, pp. 39--48.

\bibitem{liu2015dependency}
Y.~Liu, F.~Wei, S.~Li, H.~Ji, M.~Zhou, H.~Wang, A dependency-based neural
  network for relation classification, arXiv preprint arXiv:1507.04646.

\bibitem{huang2016attention}
X.~Huang, et~al., Attention-based convolutional neural network for semantic
  relation extraction, in: Proceedings of COLING 2016, the 26th International
  Conference on Computational Linguistics: Technical Papers, 2016, pp.
  2526--2536.

\bibitem{zhang2015relation}
D.~Zhang, D.~Wang, Relation classification via recurrent neural network, arXiv
  preprint arXiv:1508.01006.

\bibitem{jacovi2018understanding}
A.~Jacovi, O.~S. Shalom, Y.~Goldberg, Understanding convolutional neural
  networks for text classification, arXiv preprint arXiv:1809.08037.

\bibitem{bai2018empirical}
S.~Bai, J.~Z. Kolter, V.~Koltun, An empirical evaluation of generic
  convolutional and recurrent networks for sequence modeling, arXiv preprint
  arXiv:1803.01271.

\bibitem{johnson2014effective}
R.~Johnson, T.~Zhang, Effective use of word order for text categorization with
  convolutional neural networks, arXiv preprint arXiv:1412.1058.

\bibitem{kim2014convolutional}
Y.~Kim, \href{http://aclweb.org/anthology/D/D14/D14-1181.pdf}{Convolutional
  neural networks for sentence classification}, in: Proceedings of the 2014
  Conference on Empirical Methods in Natural Language Processing, {EMNLP} 2014,
  October 25-29, 2014, Doha, Qatar, {A} meeting of SIGDAT, a Special Interest
  Group of the {ACL}, 2014, pp. 1746--1751.
\newline\urlprefix\url{http://aclweb.org/anthology/D/D14/D14-1181.pdf}

\bibitem{wang2018copenhagen}
D.~Wang, J.~G. Simonsen, B.~Larsen, C.~Lioma, The copenhagen team participation
  in the factuality task of the competition of automatic identification and
  verification of claims in political debates of the clef-2018 fact checking
  lab., in: CLEF (Working Notes), 2018.

\bibitem{conneau2017supervised}
A.~Conneau, D.~Kiela, H.~Schwenk, L.~Barrault, A.~Bordes, Supervised learning
  of universal sentence representations from natural language inference data,
  arXiv preprint arXiv:1705.02364.

\bibitem{zhou2016attention}
P.~Zhou, W.~Shi, J.~Tian, Z.~Qi, B.~Li, H.~Hao, B.~Xu, Attention-based
  bidirectional long short-term memory networks for relation classification,
  in: Proceedings of the 54th Annual Meeting of the Association for
  Computational Linguistics (Volume 2: Short Papers), 2016, pp. 207--212.

\bibitem{zhang2018graph}
Y.~Zhang, P.~Qi, C.~D. Manning, Graph convolution over pruned dependency trees
  improves relation extraction, arXiv preprint arXiv:1809.10185.

\bibitem{socher2012semantic}
R.~Socher, B.~Huval, C.~D. Manning, A.~Y. Ng, Semantic compositionality through
  recursive matrix-vector spaces, in: Proceedings of the 2012 joint conference
  on empirical methods in natural language processing and computational natural
  language learning, Association for Computational Linguistics, 2012, pp.
  1201--1211.

\bibitem{zhang2016relation}
D.~Zhang, D.~Wang, Relation classification: Cnn or rnn?, in: Natural Language
  Understanding and Intelligent Applications, Springer, 2016, pp. 665--675.

\bibitem{devlin2018bert}
J.~Devlin, M.-W. Chang, K.~Lee, K.~Toutanova, Bert: Pre-training of deep
  bidirectional transformers for language understanding, arXiv preprint
  arXiv:1810.04805.

\end{thebibliography}

\end{document}